\pdfoutput=1
\documentclass[11pt]{article}
\usepackage{graphicx}
\usepackage{acl}
\usepackage{times}
\usepackage{latexsym}
\usepackage{graphicx}
\usepackage[T1]{fontenc}
\usepackage[utf8]{inputenc}

\usepackage{microtype}
\usepackage{lscape}
\usepackage{inconsolata}
\usepackage{threeparttable}
\usepackage{booktabs}
\usepackage{adjustbox}
\usepackage{rotating}
\usepackage{float}
\usepackage{url}
\usepackage{multirow} %
\title{SportQA: A Benchmark for Sports Understanding in Large Language Models}

\author{Author 1 \and ... \and Author n \\
         Address line \\ ... \\ Address line}
\author{Haotian Xia\textsuperscript{1},
Zhengbang Yang\textsuperscript{1},
Yuqing Wang\textsuperscript{2},
Rhys Tracy\textsuperscript{3},
Yun Zhao\textsuperscript{4}, \\
{\bf Dongdong Huang\textsuperscript{5},
Zezhi Chen\textsuperscript{1},
Yan Zhu\textsuperscript{5}$^*$,
Yuan-fang Wang\textsuperscript{3},
Weining Shen\textsuperscript{1}}$^*$ \\
\textsuperscript{1}University of California, Irvine, CA, USA \\
\textsuperscript{2}Stanford University, CA, USA\\
\textsuperscript{3}University of California, Santa Barbara, CA, USA\\
\textsuperscript{4}Meta Platforms, Inc., CA, USA\\
\textsuperscript{5} Beijing Normal University, Beijing, China\\
  \texttt{\small\{xiah6, weinings\}@uci.edu},
  \texttt{\small\{rhystracy, yfwang\}@cs.ucsb.edu},
  \texttt{\small zhuyan@bnu.edu.cn}
  }
\begin{document}
\maketitle
\def\thefootnote{*}\footnotetext{Corresponding authors.}\def\thefootnote{\arabic{footnote}}
\begin{abstract}

A deep understanding of sports, a field rich in strategic and dynamic content, is crucial for advancing Natural Language Processing (NLP). This holds particular significance in the context of evaluating and advancing Large Language Models (LLMs), given the existing gap in specialized benchmarks. To bridge this gap, we introduce SportQA, a novel benchmark specifically designed for evaluating LLMs in the context of sports understanding. SportQA encompasses over 70,000 multiple-choice questions across three distinct difficulty levels, each targeting different aspects of sports knowledge from basic historical facts to intricate, scenario-based reasoning tasks. We conducted a thorough evaluation of prevalent LLMs, mainly utilizing few-shot learning paradigms supplemented by chain-of-thought (CoT) prompting. Our results reveal that while LLMs exhibit competent performance in basic sports knowledge, they struggle with more complex, scenario-based sports reasoning, lagging behind human expertise. The introduction of SportQA marks a significant step forward in NLP, offering a tool for assessing and enhancing sports understanding in LLMs. The dataset is available at \url{https://github.com/haotianxia/SportQA} 
\end{abstract}

\section{Introduction}

The dynamic and multifaceted world of sports, characterized by its fast pace, variety of types, abundance of strategies, and rich player narratives, presents a unique set of challenges for the sports understanding capabilities of Large Language Models (LLMs). Although LLMs have shown exceptional capabilities in many Natural Language Processing (NLP) tasks such as natural language understanding (NLU)~\citep{fei-etal-2023-reasoning}, information extraction~\citep{ding-etal-2023-gpt,cong-etal-2023-universal}, and question answering (QA)~\citep{zhao-etal-2023-verify,li-etal-2023-shot}, their application in the sports domain---which involves a complex blend of statistical data, narrative content, and strategic planning---remains underexplored. Sports enthusiasts can easily answer questions like "Who won the 2022 FIFA World Cup?" or "What is the record for the most points scored in an NBA game?", but more complex queries like "Why does the float serve appear more in low-level or young-age volleyball games than in high-level games?" require expert knowledge and experience. These challenges, ranging from general knowledge to expert-level analysis, underscore the need for a dedicated sports-focused question-answering (QA) dataset to improve LLMs' comprehension and contextualization of sports information.

\begin{figure*}[h]
\centering
\includegraphics[width =1\textwidth]{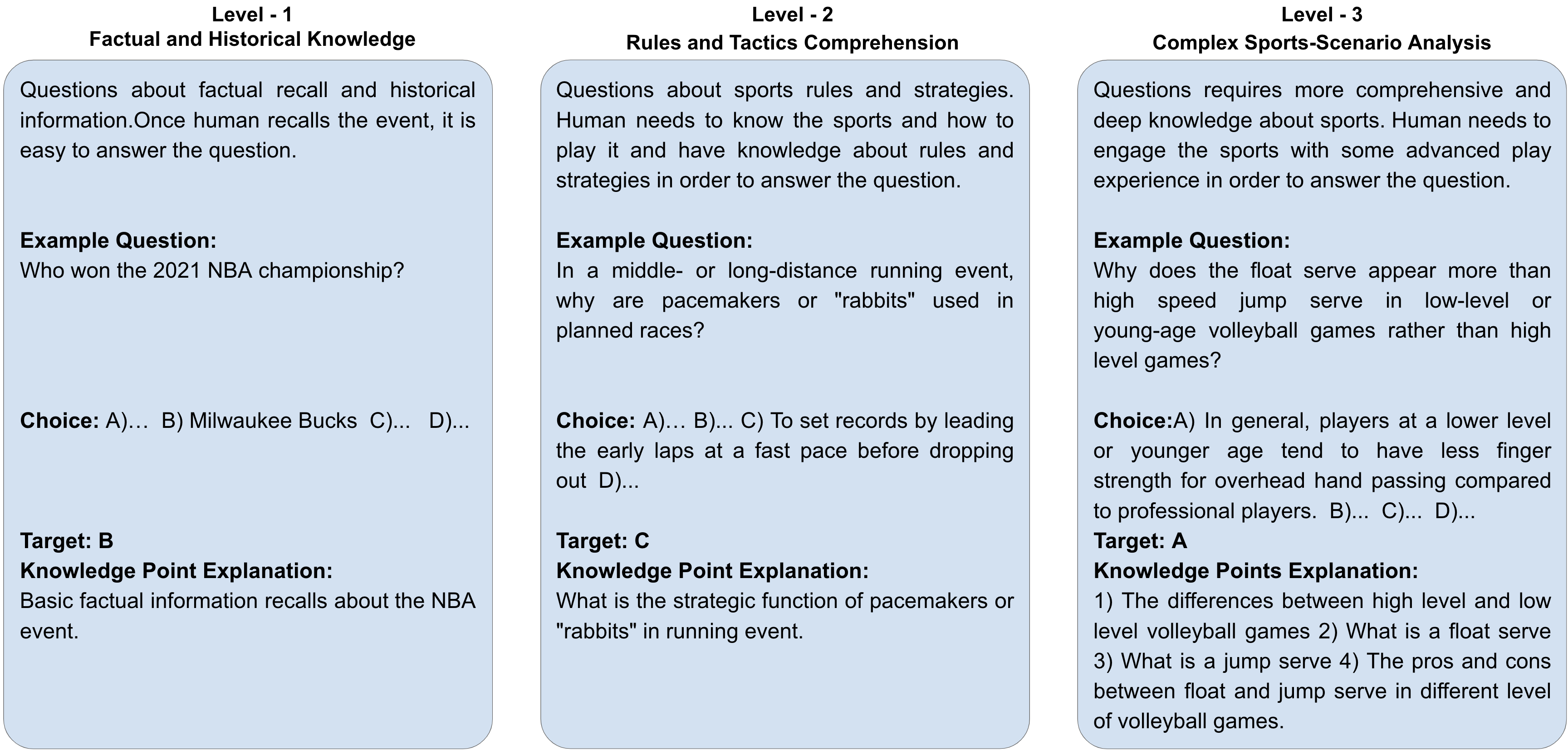}
\caption{Illustration of three levels of understanding in sports.}
\label{definition}
\end{figure*}

To evaluate the sports understanding capabilities of LLMs, the sports QA task has been introduced with several datasets for evaluation. The BIG-bench sports understanding task \citep{srivastava2023beyond} focuses on factual sports knowledge, while LiveQA \citep{liu2020liveqa} is based on Chinese live-broadcasting NBA passages. These datasets illustrate certain aspects of sports understanding. However, they exhibit limitations in size, sports domain coverage, and depth of understanding. For instance, BIG-bench might ask LLMs to judge the plausibility of a statement like “Tom Brady threw a touchdown in the Champions League Final,” merging an American football player with a soccer event. This approach, focusing more on superficial sports associations than on deep understanding of sports contexts and rules, leads to an incomplete assessment of LLMs' true sports understanding. To bridge this gap, we introduce SportQA, a comprehensive dataset crafted to challenge and evaluate LLMs in the domain of sports understanding. It consists of 70,592 multiple-choice questions, offering varying levels of difficulty, from straightforward historical facts to complex, scenario-based reasoning tasks that necessitate extensive sports knowledge and experience.

As shown in Figure~\ref{definition}, humans' understanding of sports can be divided into three levels of difficulty: factual and historical knowledge (Level-1), rules and tactics comprehension (Level-2), and complex sports-scenario analysis (Level-3). In our SportQA, level-1 understanding, covering 21,385 problems, allows humans to answer by recalling facts without needing additional sports expertise, such as knowing who won a specific Olympic event. Level-2, with 45,685 problems, suitable for sports enthusiasts, requires some expertise to understand diverse sports strategies and rules, like understanding soccer's offside rules. Level-3, encompassing 3,522 problems, is designed for sports specialists with years of experience, featuring more complex scenarios. One example under this level would be determining the best technique for a volleyball player facing three blockers during a spike in an amateur game. Level-1 and Level-2 problems are presented as multiple-choice questions with one correct answer each, while Level-3 problems follow a ‘Multiple Select’ format, allowing for one to four correct answers, and include both single and multi-hop questions in two levels of difficulty. Overall, three levels of questions collectively provide a comprehensive set to evaluate LLMs' sports understanding across a spectrum of expertise.

To gain deeper insights into the sports understanding challenges posed by SportQA, we extensively evaluated several popular recent LLMs, including Llama2~\citep{touvron2023llama}, PaLM2~\citep{anil2023palm}, GPT-3.5, and GPT-4~\citep{openai2023gpt4}. The models were assessed through few-shot standard prompting, as well as chain-of-thought prompting~\citep{wei2022chain}. Our findings show that GPT-4 outperforms in all levels, achieving an average accuracy of 82.16\% in Level-1, 75\% in Level-2, and 47.14\% in level-3. However, in Level-3, the most challenging level, GPT-4's performance is about 45\% inferior compared to human experts' accuracy in the same test questions, indicating that its understanding of sports significantly lags behind human capabilities, suggesting considerable scope for improvement in this area.

In summary, our contributions are threefold:

\begin{itemize}
    \item[(1)] We introduced SportQA, the first comprehensive dataset tailored for sports understanding in LLMs. It encompasses three levels of difficulty, aligned with human comprehension of sports. The dataset covers a diverse range of questions, from basic sports history to complex, scenario-based queries, thereby establishing itself as an essential tool for assessing the LLMs' sports understanding capabilities.
    \item[(2)] We conducted extensive experiments on SportQA, evaluating recent LLMs' sports comprehension capabilities and gained insights into their strengths and areas for improvement through manual error analysis. This research has contributed to a better understanding of LLM performance within the context of sports NLP.
    \item[(3)] We explored new directions for NLP in sports, underscoring its potential to enhance sports journalism and facilitate communication between athletes and coaches. This work not only broadens the use of NLP technologies but also lays the groundwork for the future integration of AI in sports-related fields. Furthermore, we made our datasets publicly available to benefit the research community.
\end{itemize}

\section{Related Work}

\textbf{Sports NLP.} Sports NLP is an emerging field, increasingly capturing interest due to its diverse applications, which range from sentiment analysis~\citep{10.1007/978-3-031-37940-6_27,ljajic2015sentiment} to game outcome predictions~\citep{beal2021combining,xia2022vren,oved2020predicting,tracy2023graph}, and from generating game summaries~\citep{thomson2020sportsett,huang2020generating} to augmenting sports videos with computer vision techniques~\citep{chen2022sporthesia}. Despite these advancements, a critical dimension remains underexplored: an in-depth understanding of sports in LLMs. Current applications primarily focus on analytics and do not delve into the complexities of sports understanding. Ensuring a deeper comprehension of sports in LLMs is crucial, as it can significantly broaden the scope and impact of NLP and LLM applications in the sports domain. This advancement is not only essential for enhancing current applications but also pivotal for exploring new avenues in Sports NLP.

\textbf{QA in Sports.}
    There have been several QA datasets involving testing LLMs reading and knowledge-based comprehension. Popular datasets include Trivia QA~\citep{joshi-etal-2017-triviaqa}, HotpotQA~\citep{yang-etal-2018-hotpotqa}, QUASAR~\citep{dhingra2017quasar, talmor-berant-2018-web}, KQA Pro \citep{cao-etal-2022-kqa} , and BoolQ~\citep{clark-etal-2019-boolq}. However, among these datasets, sports-related QA is scarce. When sports topics are included, they usually focus on historical facts or well-known events, rather than the intricacies of sports rules, strategies, and real-time decision-making. This focus on factual recall rather than a deeper, strategic understanding of sports limits LLMs in demonstrating nuanced comprehension of this special domain.

Recent work~\citep{wei2022chain} has examined LLMs' sports understanding using BIG-bench's dataset on sports subtask~\citep{srivastava2023beyond}, focusing on distinguishing plausible from implausible sports statements. This dataset contains 986 questions that require knowledge of both the names of athletes and actions common in particular sports. However, understanding a sport involves more than merely matching names and activities. QASports~\citep{jardim2023qasports} is a context extractive question-answering dataset focusing on soccer, American football, and basketball. It primarily involves answering questions based on specifically provided paragraphs focused on fact-based questions. Similarly, LiveQA~\citep{liu2020liveqa} challenges LLMs with basketball-related questions derived from play-by-play broadcast passages, focusing on understanding basketball from given paragraphs. Typical question types include timeline assessment, event tracking, and mathematical computations, such as ‘Which team will make a three-pointer first in the second quarter?’ While this tests comprehension within a specific context, it does not extensively cover other crucial aspects of sports understanding, such as strategy comparisons. The common drawback of these sports QA datasets is their focus on surface-level insights, such as basic facts and famous activities. True sports understanding involves rules, gameplay, and tactics. In contrast, our SportQA benchmark provides a comprehensive scope, covering a wide range of sports subjects, from facts and history to rules and complex scenarios. Moreover, our dataset is able to assess LLMs' intrinsic sports knowledge beyond just finding information in the provided paragraphs. We aim to address the limitations of existing datasets, fostering a deeper understanding of sports. 

%\begin{flushleft}
\textbf{Paradigms in LLMs Training.} Pre-trained language models on diverse texts, exemplified by BERT \citep{devlin-etal-2019-bert} has been a cornerstone in NLP. These models have been effectively used in tasks ranging from disease prediction~\citep{zhao2021empirical} to text classification~\citep{wang2022integrating}. However, the introduction of GPT-3~\citep{brown2020language} marked a paradigm shift from extensive task-specific fine-tuning to zero-shot and few-shot learning approaches, enabling adaptation to new tasks with minimal training. This transition has led to the development of advanced prompting techniques to enhance LLMs' understanding and reasoning abilities. Representative examples include chain-of-thought (CoT) prompting~\citep{wei2022chain}, zero-shot CoT prompting~\citep{kojima2022large}, self-consistency~\citep{wang2022self}, Tree-of-Thought prompting~\citep{yao2023tree}, and metacognitive prompting~\citep{wang2023metacognitive}. This evolution has necessitated a focus on the latest advancements in LLMs. Consequently, our study concentrates on evaluating recent LLMs, such as Llama2~\citep{touvron2023llama}, PaLM2~\citep{anil2023palm}, GPT-3.5, and GPT-4~\citep{openai2023gpt4} in diverse sports understanding tasks to explore their capabilities.
%\end{flushleft}

\section{Sports Understanding Benchmark}
We introduce SportQA, a benchmark specifically designed to evaluate the understanding of sports for LLMs. Developed in close collaboration with sports experts, questions in SportQA consist of three main levels of difficulty, each focusing on a distinct aspect of sports comprehension.

\noindent \textbf{Level-1: Foundational Sports Knowledge.} Questions in this level assess basic sports knowledge, focusing on factual recall and historical information. It provides a foundation for understanding the broader context of sports within LLMs.

\noindent \textbf{Level-2: Rules and Tactics Comprehension.} This level's questions mainly evaluate LLMs' understanding of sports rules and strategies. Covering 35 distinct sports types, it tests the models' ability to interpret game rules and employ strategic thinking, offering deeper insights into sports comprehension. Additionally, it includes more complex fact-based historical questions to ensure a well-rounded assessment of sports knowledge.

\noindent \textbf{level-3: Advanced Scenario-Based Understanding.} In this level, questions focus on evaluating the synthesis and integration of diverse sports knowledge. These involve complex, scenario-based questions that replicate real-world sports situations, requiring deep comprehension and advanced analytical thinking from LLMs.

Overall, the benchmark comprises a total of 70,592 questions. For each dataset and level, we provide a few-shot development set consisting of 5 questions per task, along with a separate test set for a thorough evaluation. Table~\ref{QAtable} provides a detailed overview of these levels and tasks, with more extensive details available in Appendix~\ref{sec:appendixA}. We employ accuracy as the evaluation metric across all tasks, ensuring a clear and objective assessment of LLMs' sports understanding capabilities. Specifically, we employ holistic accuracy for the Level 3 multi-hop experiment, where an entire question is marked as incorrect if the model fails to answer any component, main or sub-question, correctly.

\begin{table*}[ht!]
  \centering
     \tabcolsep 3pt
  \begin{adjustbox}{width=\textwidth}
  \begin{threeparttable}
  \begin{tabular}{@{}ccccc@{}}
  \toprule
  \textbf{Task} & \textbf{Data Size} & \textbf{Metrics} & \textbf{Answer Type} & \textbf{Text Sources}              \\ \midrule
  \multicolumn{5}{c}{\textbf{Level-1: Foundational Sports Knowledge} }                        \\ \midrule
  Integrating Existing Datasets   & 21,385    & Acc.    & 4-Way or 2-Way MC    & Trivia QA\textsuperscript{1}, QUASAR\textsuperscript{2}, Hotpot QA\textsuperscript{3}, KQA Pro\textsuperscript{4}, BoolQ\textsuperscript{5}   \\
 \midrule
  \multicolumn{5}{c}{\textbf{level-2: Rules and Tactics Comprehension}}                     \\ \midrule
    35 Sports and Their Variation Tasks  & 45,685     & Acc.    & 4-Way MC    & Wikipedia                          \\
  \midrule
  \multicolumn{5}{c}{\textbf{level-3: Scenario-based Questions}}              \\ \midrule
    6 Sports Easy Multi-hop Tasks     & 915   & Acc. & Multi-hop 4-Way MS    &\multirow{4}{*}{Experts Proposed Assessment Angles} \\
    6 Sports Hard Multi-hop Tasks    & 808   & Acc. & Multi-hop 4-Way MS   &   \\
  6 Sports Easy Single-hop Tasks & 903   & Acc.    & Single-hop 4-Way MS  &  \\
  6 Sports Hard Single-hop Tasks  & 896    & Acc.    &Single-hop 4-Way MS  &  \\
  \bottomrule
  \end{tabular}
  \begin{tablenotes}
    \footnotesize
    \item \textsuperscript{1}\citep{joshi-etal-2017-triviaqa}, \textsuperscript{2}\citep{dhingra2017quasar}, \textsuperscript{3}\citep{yang-etal-2018-hotpotqa}, \textsuperscript{4}\citep{cao-etal-2022-kqa}, \textsuperscript{5}\citep{clark-etal-2019-boolq}
  \end{tablenotes}
\end{threeparttable}
\end{adjustbox}
  \caption{Overview of tasks included in SportQA. The ``Data Size'' column aggregates totals from both the training, development and test sets. ``$K$-Way MC'' and ``$K$-Way MS'' signifies a multiple-choice response format and multiple-select choice (one or multiple correct answers) format respectively with $K$ options. Level-1 focuses on historical and factual sports knowledge. Level-2 consists of 35 sports types about rules, strategies, and facts, and Level-3 includes 6 sports with complex and multiple knowledge points questions. Detailed statistics on question distribution and topic focus within each task are available in the Appendix~\ref{sec:appendixA}.}
  \label{QAtable}
\end{table*}

\subsection{Dataset Construction and Quality Verification}
\textbf{Dataset Construction.} We constructed the SportQA dataset through a hybrid approach. For Level-1 and -2 problems, we combined automated templates with expert-driven modifications to create a diverse range of questions, ensuring both consistency and comprehensive coverage. For the more complex level-3 questions, we exclusively used manual question creation by experienced sports experts. This dual approach, integrating both automated and manual processes, was instrumental in crafting questions that accurately reflect the multifaceted and intricate aspects of sports knowledge.

\noindent \textbf{Quality Verification.} To ensure the integrity and accuracy of SportQA, a meticulous manual review process was essential due to the complexity of sports knowledge. This task was carried out by a highly skilled team of 36 intercollegiate student-athletes from both the US and China, each with a minimum of 8 years of sports training experience. Their extensive experience and expertise in various sports were crucial in evaluating the context and accuracy of each question. Their deep understanding of sports rules, strategies, and contexts ensured that every question in the dataset was not only factually accurate but also relevant and challenging.

During the recruitment phase, each potential team member was interviewed using ten examples for every level of questions and asked to compose example Q\&A pairs. After hiring, each member first goes through a training session to learn the task and the annotation process. Once they fully master the annotation process, we launch the official batches for them to work on.

\subsection{Level-1: Foundational Sports Knowledge}
Level-1 problems aim to assess LLMs' proficiency in foundational sports knowledge, focusing on factual recall and historical information. This level includes a total of 21,385 multiple-choice questions, sourced from a variety of QA datasets. As these datasets have different Q\&A formats (true/false, multiple choices, and open-ended narratives), efforts were spent to normalize the formats and check for correctness and relevance. 

Questions from Trivia QA~\citep{joshi-etal-2017-triviaqa}, QUASAR~\citep{dhingra2017quasar}, and Hotpot QA~\citep{yang-etal-2018-hotpotqa} were originally in open-answer format and have been adapted for this benchmark. For BoolQ~\citep{clark-etal-2019-boolq}, which primarily featured true/false questions, we adapted these into 2-way multiple-choice formats. KQA Pro~\citep{cao-etal-2022-kqa} questions, already in a multiple-choice format, were also included and checked for accuracy and relevance. The transformation of open-answer questions into multiple-choice format followed by automated template generation or manual refinement. 

\noindent \textbf{Step 1: Automated Template Generation.} For each answer type (sport, time, country/location, person, numbers, etc.), we utilized different semantic libraries. For example, (1) Sports: If the answer was a sport, other sports were automatically selected as distractors. (2) Time: Adjacent periods were chosen for time-based answers. (3) Country/Location: Similar geographical locations were used as distractors. (4) Person: A specialized semantic library of names was created for each sport to gather information about athletes, and this library was then used to select distractors.

\noindent \textbf{Step 2: Manual Refinement.} Questions requiring more nuanced distractors or falling outside these categories were further refined by our review team of 36 student-athletes, who manually crafted additional misleading options.

Level-1 questions establish the foundation for sports knowledge. However, due to the limited scope and specificity of sports in existing datasets, level-2 is introduced to expand this scope. It addresses the need for professional knowledge of sports, such as rules and strategies.

\subsection{level-2: Rules and Tactics Comprehension}

level-2 problems are designed to conduct an in-depth assessment of LLMs' understanding of sports rules, tactics, and an extended range of historical and factual knowledge. This level comprises 45,685 questions, covering a broad spectrum of sports disciplines. 

\noindent \textbf{Step 1: Content Categorization and Annotation.} The review team categorized content from Wikipedia related to 35 distinct sports tasks, including the 28 Olympic sports and their variation, the four new sports (Breaking, sport climbing, skateboarding, and surfing) debuting in the 2024 Paris Olympics, and popular non-Olympic sports like baseball, ice hockey, and American football. The focus was on compiling sources primarily about rules and tactics, supplemented with historical and factual content for comprehensive coverage. They performed categorization and annotation of content from Wikipedia, identifying and annotating critical information points within each sport's rules, tactics, and historical contexts.

\noindent \textbf{Step 2: Hybrid Question Development and Templates.} Our approach to question development was twofold. First, we employed a set of predefined templates to ensure consistent coverage across various sports, and examples of these templates are illustrated in Appendix \ref{sec:appendixA}. Furthermore, for scenarios that required deeper exploration beyond these templates, we manually crafted questions, ensuring a rich and diverse set of queries based on the annotation of content and knowledge points.

\noindent \textbf{Step 3: Distractor Generation and Categorization.} The distractor generation for SportQA was a nuanced process. Automated distractors for template-based questions were created using a specialized, sport-specific library, ensuring each was relevant and challenging. This library, organized by sport, included diverse categories such as athletes' names and tactics. For questions crafted manually, our team of experts meticulously developed plausible yet incorrect distractors, adding depth and complexity to the dataset.

\noindent \textbf{Step 4: Expert Review and Alignment.} The review team ensured each question's alignment with the source content, verifying the efficacy of questions and eliminating outdated or irrelevant information.

By following the above steps, level-2 problems not only offer a comprehensive evaluation of sports rules and tactics but also enrich the dataset with an expansive array of sports history and factual knowledge, enhancing the overall utility of SportQA.

\subsection{level-3: Scenario-Based Questions}
level-3 problems are a comprehensive and nuanced component of our benchmark, comprising 3,522 scenario-based questions across six key sports. This level includes football (soccer), basketball, volleyball, tennis, table tennis, and American football. Each sport features a set of both multi-hop and single-hop questions with multiple-select types (one to four correct answers), meticulously categorized into easy and hard difficulty levels. The categorization is based on the complexity of the scenarios and the number of integrated knowledge points required to answer each question. This results in four distinct tasks per sport, 24 tasks in total, closely mirroring the intricacies and complexities of real-world sports situations. 

The requirement for manual question creation in level-3 is rooted in the sophisticated and deep understanding of each sport needed to craft these questions. This level's complexity demands not just a surface-level acquaintance with the sports but an in-depth, experiential knowledge, often only possessed by those who have actively participated or been deeply involved in the sport. Such profound insight is crucial for developing questions that accurately capture the nuances, strategic elements, and practical applications inherent in real-world sports scenarios. This elevates level-3 above Levels 1 and 2, making it a more advanced and comprehensive test of sports understanding and analysis.

\noindent \textbf{Step 1: Expert Proposed Assessment Angle.} We first invited coaches to propose the assessment angles, which are further detailed in Appendix~\ref{sec:appendixA}, for these sports based on their expertise. Their extensive coaching experience and mastery understanding of sports ensure that the knowledge points behind each question are meaningful and effective for the sports.

\noindent \textbf{Step 2: Manual Questions Generation.} After coaches proposed the assessment angles across all 6 sports, these angles were used by our review team, combining their training and competition experience with their unique advanced understanding and expertise in sports, to develop questions. This ensures a comprehensive assessment of LLMs' sports understanding and analytical skills.

The distribution of questions per sport, shown in Appendix~\ref{sec:appendixA}, is directly influenced by the availability and expertise of our review team. Sports with more expert reviewers are represented with a larger number of questions, ensuring the authenticity and depth of the content. This alignment guarantees that the questions are not only challenging but also reflect the dynamic nature of each sport.

The manual creation process of level-3 emphasizes the professional rigor and subject-matter expertise that underpin SportQA, making it an exceptional measure of sports knowledge.

\begin{table*}[ht!]
\begin{adjustbox}{width=\textwidth}
\centering
\begin{tabular}{lc c c c c cc}
\hline
\textbf{Model} & \textbf{Level-1} & \textbf{level-2} & \textbf{level-3 Easy Single-hop} & \textbf{level-3 Hard Single-hop} & \textbf{level-3 Easy Multi-hop} & \textbf{level-3 Hard Multi-hop} \\ \hline
Llama2-13b(0S,CoT) & 50.90 & 52.32 & 21.46 & 15.16 & 14.80 & 9.20  \\
Llama2-13b(5S,SP) & 42.75 & 48.02 & 25.10 & 25.82 & 5.60 & 5.86 \\
Llama2-13b(5S,CoT) & 48.65 & 51.54 & 26.72 & 32.38 & 9.20 & 8.79\\  \hline
PaLM2(0S,CoT) & 59.35 & 48.28 & 48.37 & 44.49 & 23.20 & 12.97  \\
PaLM2(5S,SP) & 64.85 & 56.62 & 49.19 & 49.80 & 29.20 & 16.74  \\
PaLM2(5S,CoT) & 66.20 & 57.02 & 47.56 & 38.37 & 19.60 & 11.29  \\ \hline
GPT3.5(0S,CoT) & 49.04 & 51.04 & 49.18 & 45.71 & 27.6 & 14.64  \\
GPT3.5(5S,SP) & 74.74 & 68.07 & 45.52 & 36.73 & 25.20 & 19.24  \\
GPT3.5(5S,CoT) & 71.97 & 66.52 & 46.34 & 45.30 & 23.60 & 14.64  \\ \hline
GPT-4(0S,CoT) & 80.60 & 69.01 & 67.07	& 55.10&	32.00	&22.59 \\
GPT-4(5S,SP) & 80.24 & 77.17 & 70.73	& 63.27	& 33.60	& \textbf{24.69} \\
GPT-4(5S,CoT) & \textbf{85.63} & \textbf{78.82} &\textbf{73.58}	&\textbf{64.08}&	\textbf{34.40}&	23.01\\ \hline
Human & - & - & 96.63 & 96.02 & 94.90 & 91.84 \\ \hline
\end{tabular} 
\end{adjustbox}
\caption{Performance comparison of each model across three levels in SportQA. GPT-4 consistently
outperforms other models under both zero-shot (0S) and 5-shot (5S) settings across all levels. Human performance serves as an upper bound, illustrating that there still exists room for improvement in LLMs on sports understanding tasks.}
\label{table:performance_comparison}
\end{table*}

\section{Experiments}
This section evaluates the performance of prevalent LLMs on our SportQA benchmark. We aim to assess the efficacy of these models in understanding complex sports scenarios across multiple tasks. We present the optimal result, derived from multiple iterations for each experimental condition.

\subsection{Experimental Setup}
We evaluated several leading language models on the SportQA benchmark, including open-source models like Llama2-13b-chat~\citep{touvron2023llama}, and closed-source models such as PaLM-bison-chat~\citep{anil2023palm}, GPT-3.5-turbo, and GPT-4~\citep{openai2023gpt4}. Access to these models was facilitated through their respective APIs. 

For Level-1, we randomly selected 2000 questions from the test set. For level-2, our sampling strategy varied based on the number of questions per sport: 30\% of all questions for sports with fewer than 200 questions, 15\% for 200-800, 5\% for 800-1500, 2.5\% for 2500-10,000, and 1.5\% for more than 10,000 questions, totaling 2243 questions. In level-3, the test sample size was determined based on the total questions of sports, with 20\% for soccer, basketball, and tennis, 30\% for volleyball, and 50\% for table tennis and American football, a total of 980 questions.

In our study, we primarily focused on the CoT prompting method for model evaluation. This decision was informed by the evidence~\citep{wei2022chain}, which highlights the effectiveness of few-shot CoT in sports understanding contexts. The CoT approach, involving a step-by-step reasoning process, is particularly suitable for complex sports understanding tasks. We also considered the zero-shot CoT method~\citep{kojima2022large} and few-shot standard prompting (SP)~\citep{brown2020language} as additional prompting baselines for evaluation. In the few-shot setting (CoT or SP), we take 5 exemplars for each task, where the answers to these demonstrations are annotated by human experts. Exemplars were consistently drawn from the development set, with the temperature parameter set to 0 to ensure consistent response generation. Additional details about the prompts we used are available in Appendix~\ref{sec:appendixB}. 

In addition to model performance evaluation, we engaged student-athletes not part of the review team, each specializing in one of the sports covered in level-3. These athletes were tasked with manually answering the test set of level-3. This effort, involving each expert in their respective sport, was undertaken to better understand human performance on these complex, scenario-based questions. Their contributions provide valuable insights into the realistic expectations of human expertise in sports understanding and serve as a standard for expert-level accuracy.

\subsection{Overall Performance Comparison}
We compared the performance of different models across all tasks in three different levels, and the overall result is shown in Table~\ref{table:performance_comparison}. Results for individual tasks are shown in Appendix~\ref{sec:appendixC}. There are several key takeaways. First, GPT-4 consistently outperforms other models across all tasks, demonstrating a performance advantage of over 15\% compared to other models on average. Regarding prompting effectiveness, we note that CoT often results in performance enhancements, which corroborates the findings from~\cite{wei2022chain}, emphasizing the efficacy of step-by-step prompting with few exemplars in augmenting LLMs’ performance in intricate reasoning tasks. Each model, including GPT-4, consistently exhibited the best performance in Level-1, with a gradual decline in accuracy moving to Levels 2 and 3. This trend aligns with the increasing complexity and sophistication designed for each successive level of our benchmark. Level-1, focusing on foundational knowledge, proved to be the most accessible, while level-3, with its advanced scenario-based tasks, posed the greatest challenge. 

Finally, while GPT-4 leads among all the models, human expertise still exceeds it by
roughly 30\% - 65\% in different tasks of level-3, highlighting the complexity of these sports understanding tasks and indicating room for future improvements in LLMs.

\begin{figure*}[htbp]
\centering
\includegraphics[width =1\textwidth]{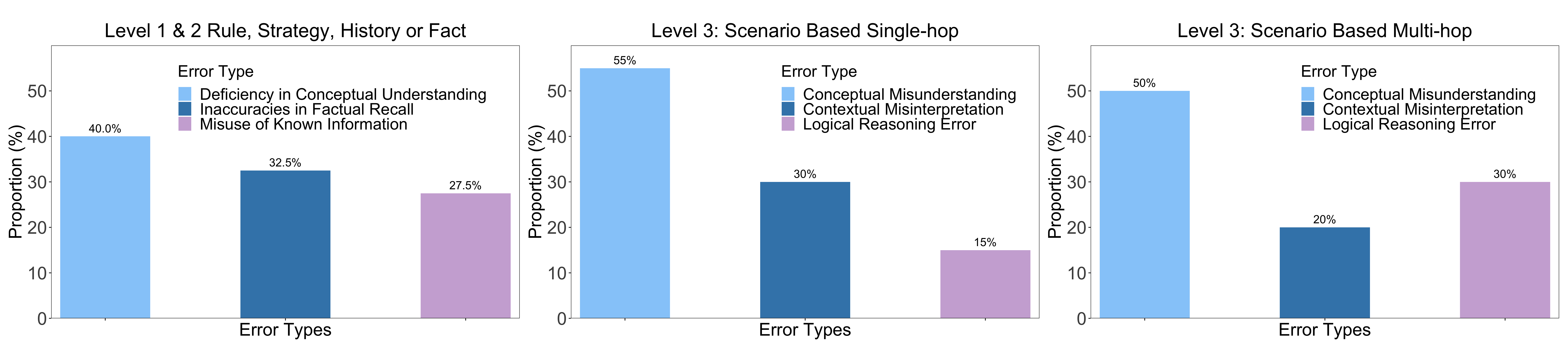}
\caption{Error type distributions across different levels of SportQA analysis, with level-3 separated into 'Single-Hop' and 'Multi-Hop' reasoning to highlight specific error trends. In levels 1 and 2, the model often struggles with the details of concepts, and these issues become amplified in level-3. This amplification of errors leads to more significant problems, resulting in worse performance of models in the more complex scenarios of level-3.}
\label{error}
\end{figure*}

\subsection{Error Analysis}

To better understand the mistakes made by models, we randomly chose and manually analyzed 20 instances from each level where a model made an incorrect choice. We prompted the model to explain its decisions, then reviewed these explanations to identify errors, understand the reasons behind them, and categorize them into specific error types. The error analysis for Levels 1 and 2 in SportQA is closely related to the fundamental nature of these levels, which focus on factual knowledge and basic rule comprehension. We identify the following error types which align with the core challenges of recalling specific sports facts, understanding basic rules, and applying simple strategies: Deficiency in Conceptual Understanding, Misuse of Known Information, and Inaccuracies in Factual Recall.

For level-3 of SportQA, given its unique question types involving scenario-based, multi-knowledge point questions with both single and multi-hop reasoning, we devised specific error categories:
Conceptual Misunderstanding, Logical Reasoning Error, and Contextual Misinterpretation.

For this analysis, we focused on GPT-4. Figure~\ref{error} shows the error types and their respective proportions for each level group. Within level-1 and level-2, “deficiency in conceptual understanding ” was the most frequent error, accounting for 40\% of all mistakes. 
In level-3 of the SportQA analysis, During the analysis process, we noticed that 82.5\% of incorrect questions have inadequacy in multifaceted answer identification. Within this, the greatest error category appears to be conceptual misunderstanding, prevalent in both single-hop and multi-hop scenarios, constituting 55\% and 50\% of the errors, respectively. For instance, there is an error in distinguishing the difference between "referee" and "mediator" in the scenario-based questions. This illustrates the model often fails to grasp the concepts that are combined with scenarios necessary for these complex questions. Detailed descriptions of each error type, along with examples and case studies of CoT incorrect outputs and their error analysis, are available in Appendix~\ref{sec:appendixD}.  

\section{Conclusion}
This study has introduced SportQA, a benchmark for assessing the understanding of sports in LLMs. Unlike previous benchmarks that primarily focused on basic fact-recall or simplistic sports-related queries, SportQA delves deeper into the intricacies of sports knowledge, spanning historical facts, rules, strategies, and scenario-based reasoning. Our evaluations reveal that while current LLMs like GPT-4 show promising capabilities in foundational knowledge and rule comprehension, their performance in complex, scenario-based reasoning remains a challenge and still lags behind human expertise. The results underscore the need for ongoing advancements in NLP and AI to achieve a deeper and more nuanced understanding of sports. Improvements to logical reasoning and in-depth understanding in the context of sports will be a gateway toward allowing Large Language Models to improve their performance and adaptability across a large range of real-world, diverse, and ever-changing domains. SportQA serves as an essential tool for future research in this area, offering a structured framework to measure and enhance LLMs' capabilities in sports understanding. 

\section*{Limitations}
While SportQA presents a broad assessment of sports understanding, we acknowledge its limitations. The primary limitation lies in the complexity of creating scenario-based questions for level-3. Due to the intricate nature and high standards required for these questions, the volume of questions and the range of sports it covers at this level is limited compared to other levels. We are dedicated to continually updating and expanding the dataset to enhance its scope and depth.

Another significant limitation of SportQA is its focus predominantly on rules and gameplay aspects of sports understanding. Critical elements such as sports medicine and psychology, which are integral to a comprehensive understanding of sports, are not currently covered in the benchmark. These areas require specialized medical and psychological knowledge, making their integration into the benchmark complex and demanding. This gap in content underscores the need for a more interdisciplinary approach in future iterations of SportQA, to encompass a wider spectrum of sports knowledge and understanding and the need to recruit team members with a more diverse background.

Furthermore, while we aimed to include a variety of LLMs in our analysis, budgetary constraints limited our ability to evaluate certain high-capacity models like the open-source Llama2-70b-chat~\citep{touvron2023llama}. This represents a limitation in our current study, as including a wider range of models could potentially offer more insights into their respective capabilities in sports understanding tasks. In future work, we plan to extend our model evaluation to encompass a broader spectrum of LLMs, ensuring a more comprehensive analysis.

\section*{Acknowledgements}

We would like to thank Weixi Feng for his helpful discussion. Moreover, we
thank anonymous reviewers for their valuable suggestions.

\bibliography{anthology,custom}

\begin{thebibliography}{35}
\expandafter\ifx\csname natexlab\endcsname\relax\def\natexlab#1{#1}\fi

\bibitem[{Anil et~al.(2023)Anil, Dai, Firat, Johnson, Lepikhin, Passos, Shakeri, Taropa, Bailey, Chen et~al.}]{anil2023palm}
Rohan Anil, Andrew~M Dai, Orhan Firat, Melvin Johnson, Dmitry Lepikhin, Alexandre Passos, Siamak Shakeri, Emanuel Taropa, Paige Bailey, Zhifeng Chen, et~al. 2023.
\newblock Palm 2 technical report.
\newblock \emph{arXiv preprint arXiv:2305.10403}.

\bibitem[{Baca et~al.(2023)Baca, Ardiles, Cruz, Mamani, and Capcha}]{10.1007/978-3-031-37940-6_27}
Luis Baca, N{\'a}tali Ardiles, Jose Cruz, Wilson Mamani, and John Capcha. 2023.
\newblock Deep learning model based on a transformers network for sentiment analysis using nlp in sports worldwide.
\newblock In \emph{Advances in Computing and Data Sciences}, pages 328--339. Springer Nature Switzerland.

\bibitem[{Beal et~al.(2021)Beal, Middleton, Norman, and Ramchurn}]{beal2021combining}
Ryan Beal, Stuart~E Middleton, Timothy~J Norman, and Sarvapali~D Ramchurn. 2021.
\newblock Combining machine learning and human experts to predict match outcomes in football: A baseline model.
\newblock In \emph{Proceedings of the AAAI Conference on Artificial Intelligence}, volume~35, pages 15447--15451.

\bibitem[{bench authors(2023)}]{srivastava2023beyond}
BIG bench authors. 2023.
\newblock \href {https://openreview.net/forum?id=uyTL5Bvosj} {Beyond the imitation game: Quantifying and extrapolating the capabilities of language models}.
\newblock \emph{Transactions on Machine Learning Research}.

\bibitem[{Brown et~al.(2020)Brown, Mann, Ryder, Subbiah, Kaplan, Dhariwal, Neelakantan, Shyam, Sastry, Askell et~al.}]{brown2020language}
Tom Brown, Benjamin Mann, Nick Ryder, Melanie Subbiah, Jared~D Kaplan, Prafulla Dhariwal, Arvind Neelakantan, Pranav Shyam, Girish Sastry, Amanda Askell, et~al. 2020.
\newblock Language models are few-shot learners.
\newblock \emph{Advances in neural information processing systems}, 33:1877--1901.

\bibitem[{Cao et~al.(2022)Cao, Shi, Pan, Nie, Xiang, Hou, Li, He, and Zhang}]{cao-etal-2022-kqa}
Shulin Cao, Jiaxin Shi, Liangming Pan, Lunyiu Nie, Yutong Xiang, Lei Hou, Juanzi Li, Bin He, and Hanwang Zhang. 2022.
\newblock \href {https://doi.org/10.18653/v1/2022.acl-long.422} {{KQA} pro: A dataset with explicit compositional programs for complex question answering over knowledge base}.
\newblock In \emph{Proceedings of the 60th Annual Meeting of the Association for Computational Linguistics (Volume 1: Long Papers)}, pages 6101--6119, Dublin, Ireland. Association for Computational Linguistics.

\bibitem[{Chen et~al.(2022)Chen, Yang, Xie, Beyer, Xia, Wu, and Pfister}]{chen2022sporthesia}
Zhutian Chen, Qisen Yang, Xiao Xie, Johanna Beyer, Haijun Xia, Yingcai Wu, and Hanspeter Pfister. 2022.
\newblock Sporthesia: Augmenting sports videos using natural language.
\newblock \emph{IEEE transactions on visualization and computer graphics}, 29(1):918--928.

\bibitem[{Clark et~al.(2019)Clark, Lee, Chang, Kwiatkowski, Collins, and Toutanova}]{clark-etal-2019-boolq}
Christopher Clark, Kenton Lee, Ming-Wei Chang, Tom Kwiatkowski, Michael Collins, and Kristina Toutanova. 2019.
\newblock \href {https://doi.org/10.18653/v1/N19-1300} {{B}ool{Q}: Exploring the surprising difficulty of natural yes/no questions}.
\newblock In \emph{Proceedings of the 2019 Conference of the North {A}merican Chapter of the Association for Computational Linguistics: Human Language Technologies, Volume 1 (Long and Short Papers)}, pages 2924--2936, Minneapolis, Minnesota. Association for Computational Linguistics.

\bibitem[{Cong et~al.(2023)Cong, Yu, Fang, Liu, Yu, Hu, Huang, Li, and Wang}]{cong-etal-2023-universal}
Xin Cong, Bowen Yu, Mengcheng Fang, Tingwen Liu, Haiyang Yu, Zhongkai Hu, Fei Huang, Yongbin Li, and Bin Wang. 2023.
\newblock \href {https://doi.org/10.18653/v1/2023.findings-acl.251} {Universal information extraction with meta-pretrained self-retrieval}.
\newblock In \emph{Findings of the Association for Computational Linguistics: ACL 2023}, pages 4084--4100, Toronto, Canada. Association for Computational Linguistics.

\bibitem[{Devlin et~al.(2019)Devlin, Chang, Lee, and Toutanova}]{devlin-etal-2019-bert}
Jacob Devlin, Ming-Wei Chang, Kenton Lee, and Kristina Toutanova. 2019.
\newblock \href {https://doi.org/10.18653/v1/N19-1423} {{BERT}: Pre-training of deep bidirectional transformers for language understanding}.
\newblock In \emph{Proceedings of the 2019 Conference of the North {A}merican Chapter of the Association for Computational Linguistics: Human Language Technologies, Volume 1 (Long and Short Papers)}, pages 4171--4186, Minneapolis, Minnesota. Association for Computational Linguistics.

\bibitem[{Dhingra et~al.(2017)Dhingra, Mazaitis, and Cohen}]{dhingra2017quasar}
Bhuwan Dhingra, Kathryn Mazaitis, and William~W. Cohen. 2017.
\newblock \href {http://arxiv.org/abs/1707.03904} {Quasar: Datasets for question answering by search and reading}.

\bibitem[{Ding et~al.(2023)Ding, Qin, Liu, Chia, Li, Joty, and Bing}]{ding-etal-2023-gpt}
Bosheng Ding, Chengwei Qin, Linlin Liu, Yew~Ken Chia, Boyang Li, Shafiq Joty, and Lidong Bing. 2023.
\newblock \href {https://doi.org/10.18653/v1/2023.acl-long.626} {Is {GPT}-3 a good data annotator?}
\newblock In \emph{Proceedings of the 61st Annual Meeting of the Association for Computational Linguistics (Volume 1: Long Papers)}, pages 11173--11195, Toronto, Canada. Association for Computational Linguistics.

\bibitem[{Fei et~al.(2023)Fei, Li, Liu, Bing, Li, and Chua}]{fei-etal-2023-reasoning}
Hao Fei, Bobo Li, Qian Liu, Lidong Bing, Fei Li, and Tat-Seng Chua. 2023.
\newblock \href {https://doi.org/10.18653/v1/2023.acl-short.101} {Reasoning implicit sentiment with chain-of-thought prompting}.
\newblock In \emph{Proceedings of the 61st Annual Meeting of the Association for Computational Linguistics (Volume 2: Short Papers)}, pages 1171--1182, Toronto, Canada. Association for Computational Linguistics.

\bibitem[{Huang et~al.(2020)Huang, Li, and Chang}]{huang2020generating}
Kuan-Hao Huang, Chen Li, and Kai-Wei Chang. 2020.
\newblock Generating sports news from live commentary: A chinese dataset for sports game summarization.
\newblock In \emph{Proceedings of the 1st Conference of the Asia-Pacific Chapter of the Association for Computational Linguistics and the 10th International Joint Conference on Natural Language Processing}, pages 609--615.

\bibitem[{Jardim et~al.(2023)Jardim, Moraes, and Aguiar}]{jardim2023qasports}
Pedro~Calciolari Jardim, Leonardo Mauro~Pereira Moraes, and Cristina~Dutra Aguiar. 2023.
\newblock Qasports: A question answering dataset about sports.
\newblock In \emph{Anais do V Dataset Showcase Workshop}, pages 1--12. SBC.

\bibitem[{Joshi et~al.(2017)Joshi, Choi, Weld, and Zettlemoyer}]{joshi-etal-2017-triviaqa}
Mandar Joshi, Eunsol Choi, Daniel Weld, and Luke Zettlemoyer. 2017.
\newblock \href {https://doi.org/10.18653/v1/P17-1147} {{T}rivia{QA}: A large scale distantly supervised challenge dataset for reading comprehension}.
\newblock In \emph{Proceedings of the 55th Annual Meeting of the Association for Computational Linguistics (Volume 1: Long Papers)}, pages 1601--1611, Vancouver, Canada. Association for Computational Linguistics.

\bibitem[{Kojima et~al.(2022)Kojima, Gu, Reid, Matsuo, and Iwasawa}]{kojima2022large}
Takeshi Kojima, Shixiang~Shane Gu, Machel Reid, Yutaka Matsuo, and Yusuke Iwasawa. 2022.
\newblock Large language models are zero-shot reasoners.
\newblock \emph{Advances in neural information processing systems}, 35:22199--22213.

\bibitem[{Li et~al.(2023)Li, Ma, Zhuang, Gu, Su, and Chen}]{li-etal-2023-shot}
Tianle Li, Xueguang Ma, Alex Zhuang, Yu~Gu, Yu~Su, and Wenhu Chen. 2023.
\newblock \href {https://doi.org/10.18653/v1/2023.acl-long.385} {Few-shot in-context learning on knowledge base question answering}.
\newblock In \emph{Proceedings of the 61st Annual Meeting of the Association for Computational Linguistics (Volume 1: Long Papers)}, pages 6966--6980, Toronto, Canada. Association for Computational Linguistics.

\bibitem[{Liu et~al.(2020)Liu, Jiang, Wang, and Li}]{liu2020liveqa}
Qianying Liu, Sicong Jiang, Yizhong Wang, and Sujian Li. 2020.
\newblock Liveqa: A question answering dataset over sports live.
\newblock In \emph{Proceedings of the 19th Chinese National Conference on Computational Linguistics}, pages 1057--1067.

\bibitem[{Ljaji{\'c} et~al.(2015)Ljaji{\'c}, Ljaji{\'c}, Spalevi{\'c}, Arsi{\'c}, and Vu{\v{c}}kovi{\'c}}]{ljajic2015sentiment}
Adela Ljaji{\'c}, Ertan Ljaji{\'c}, Petar Spalevi{\'c}, Branko Arsi{\'c}, and Darko Vu{\v{c}}kovi{\'c}. 2015.
\newblock Sentiment analysis of textual comments in field of sport.
\newblock In \emph{24nd International Electrotechnical and Computer Science Conference (ERK 2015), IEEE, Slovenia}.

\bibitem[{OpenAI(2023)}]{openai2023gpt4}
OpenAI. 2023.
\newblock \href {http://arxiv.org/abs/2303.08774} {Gpt-4 technical report}.

\bibitem[{Oved et~al.(2020)Oved, Feder, and Reichart}]{oved2020predicting}
Nadav Oved, Amir Feder, and Roi Reichart. 2020.
\newblock Predicting in-game actions from interviews of nba players.
\newblock \emph{Computational Linguistics}, 46(3):667--712.

\bibitem[{Talmor and Berant(2018)}]{talmor-berant-2018-web}
Alon Talmor and Jonathan Berant. 2018.
\newblock \href {https://doi.org/10.18653/v1/N18-1059} {The web as a knowledge-base for answering complex questions}.
\newblock In \emph{Proceedings of the 2018 Conference of the North {A}merican Chapter of the Association for Computational Linguistics: Human Language Technologies, Volume 1 (Long Papers)}, pages 641--651, New Orleans, Louisiana. Association for Computational Linguistics.

\bibitem[{Thomson et~al.(2020)Thomson, Reiter, and Sripada}]{thomson2020sportsett}
Craig Thomson, Ehud Reiter, and Somayajulu Sripada. 2020.
\newblock Sportsett: basketball-a robust and maintainable data-set for natural language generation.
\newblock In \emph{Proceedings of the Workshop on Intelligent Information Processing and Natural Language Generation}, pages 32--40.

\bibitem[{Touvron et~al.(2023)Touvron, Martin, Stone, Albert, Almahairi, Babaei, Bashlykov, Batra, Bhargava, Bhosale et~al.}]{touvron2023llama}
Hugo Touvron, Louis Martin, Kevin Stone, Peter Albert, Amjad Almahairi, Yasmine Babaei, Nikolay Bashlykov, Soumya Batra, Prajjwal Bhargava, Shruti Bhosale, et~al. 2023.
\newblock Llama 2: Open foundation and fine-tuned chat models.
\newblock \emph{arXiv preprint arXiv:2307.09288}.

\bibitem[{Tracy et~al.(2023)Tracy, Xia, Rasla, Wang, and Singh}]{tracy2023graph}
Rhys Tracy, Haotian Xia, Alex Rasla, Yuan-Fang Wang, and Ambuj Singh. 2023.
\newblock Graph encoding and neural network approaches for volleyball analytics: From game outcome to individual play predictions.
\newblock \emph{arXiv preprint arXiv:2308.11142}.

\bibitem[{Wang et~al.(2022{\natexlab{a}})Wang, Wei, Schuurmans, Le, Chi, Narang, Chowdhery, and Zhou}]{wang2022self}
Xuezhi Wang, Jason Wei, Dale Schuurmans, Quoc Le, Ed~Chi, Sharan Narang, Aakanksha Chowdhery, and Denny Zhou. 2022{\natexlab{a}}.
\newblock Self-consistency improves chain of thought reasoning in language models.
\newblock \emph{arXiv preprint arXiv:2203.11171}.

\bibitem[{Wang and Zhao(2023)}]{wang2023metacognitive}
Yuqing Wang and Yun Zhao. 2023.
\newblock Metacognitive prompting improves understanding in large language models.
\newblock \emph{arXiv preprint arXiv:2308.05342}.

\bibitem[{Wang et~al.(2022{\natexlab{b}})Wang, Zhao, Callcut, and Petzold}]{wang2022integrating}
Yuqing Wang, Yun Zhao, Rachael Callcut, and Linda Petzold. 2022{\natexlab{b}}.
\newblock Integrating physiological time series and clinical notes with transformer for early prediction of sepsis.
\newblock \emph{arXiv preprint arXiv:2203.14469}.

\bibitem[{Wei et~al.(2022)Wei, Wang, Schuurmans, Bosma, Xia, Chi, Le, Zhou et~al.}]{wei2022chain}
Jason Wei, Xuezhi Wang, Dale Schuurmans, Maarten Bosma, Fei Xia, Ed~Chi, Quoc~V Le, Denny Zhou, et~al. 2022.
\newblock Chain-of-thought prompting elicits reasoning in large language models.
\newblock \emph{Advances in Neural Information Processing Systems}, 35:24824--24837.

\bibitem[{Xia et~al.(2022)Xia, Tracy, Zhao, Fraisse, Wang, and Petzold}]{xia2022vren}
Haotian Xia, Rhys Tracy, Yun Zhao, Erwan Fraisse, Yuan-Fang Wang, and Linda Petzold. 2022.
\newblock Vren: Volleyball rally dataset with expression notation language.
\newblock In \emph{2022 IEEE International Conference on Knowledge Graph (ICKG)}, pages 337--346. IEEE.

\bibitem[{Yang et~al.(2018)Yang, Qi, Zhang, Bengio, Cohen, Salakhutdinov, and Manning}]{yang-etal-2018-hotpotqa}
Zhilin Yang, Peng Qi, Saizheng Zhang, Yoshua Bengio, William Cohen, Ruslan Salakhutdinov, and Christopher~D. Manning. 2018.
\newblock \href {https://doi.org/10.18653/v1/D18-1259} {{H}otpot{QA}: A dataset for diverse, explainable multi-hop question answering}.
\newblock In \emph{Proceedings of the 2018 Conference on Empirical Methods in Natural Language Processing}, pages 2369--2380, Brussels, Belgium. Association for Computational Linguistics.

\bibitem[{Yao et~al.(2023)Yao, Yu, Zhao, Shafran, Griffiths, Cao, and Narasimhan}]{yao2023tree}
Shunyu Yao, Dian Yu, Jeffrey Zhao, Izhak Shafran, Thomas~L Griffiths, Yuan Cao, and Karthik Narasimhan. 2023.
\newblock Tree of thoughts: Deliberate problem solving with large language models.
\newblock \emph{arXiv preprint arXiv:2305.10601}.

\bibitem[{Zhao et~al.(2023)Zhao, Li, Joty, Qin, and Bing}]{zhao-etal-2023-verify}
Ruochen Zhao, Xingxuan Li, Shafiq Joty, Chengwei Qin, and Lidong Bing. 2023.
\newblock \href {https://doi.org/10.18653/v1/2023.acl-long.320} {Verify-and-edit: A knowledge-enhanced chain-of-thought framework}.
\newblock In \emph{Proceedings of the 61st Annual Meeting of the Association for Computational Linguistics (Volume 1: Long Papers)}, pages 5823--5840, Toronto, Canada. Association for Computational Linguistics.

\bibitem[{Zhao et~al.(2021)Zhao, Wang, Liu, Xia, Xu, Hong, Zhou, and Petzold}]{zhao2021empirical}
Yun Zhao, Yuqing Wang, Junfeng Liu, Haotian Xia, Zhenni Xu, Qinghang Hong, Zhiyang Zhou, and Linda Petzold. 2021.
\newblock Empirical quantitative analysis of covid-19 forecasting models.
\newblock In \emph{2021 International Conference on Data Mining Workshops (ICDMW)}, pages 517--526. IEEE.

\end{thebibliography}

\appendix

\section{Detailed Question Distribution and Sample Questions for SportQA}
\label{sec:appendixA}

This appendix provides an in-depth look at the question distribution across different sports and question types within SportQA, as well as illustrative examples of the types of questions featured in the dataset.

\begin{figure}[ht]
\centering
\caption{Comparative Analysis of 'Rule or Strategy' vs. 'History or Fact'}
\includegraphics[width=\linewidth]{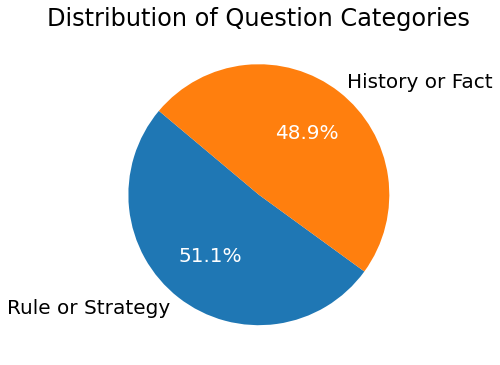}
\label{pie_chart}
\end{figure}

 \subsection{level-2 Question Distribution Across Sports}
\begin{figure*}[ht]
\centering
\caption{Proportional Distribution of Question Categories Across Sports. This bar chart illustrates the comparative distribution of 'Rule or Strategy' and 'History or Fact' questions in various sports, highlighting the dominant category in each sport and indicating the variability in focus across different sports disciplines.}
 \includegraphics[angle=270,origin=c,width=\linewidth]{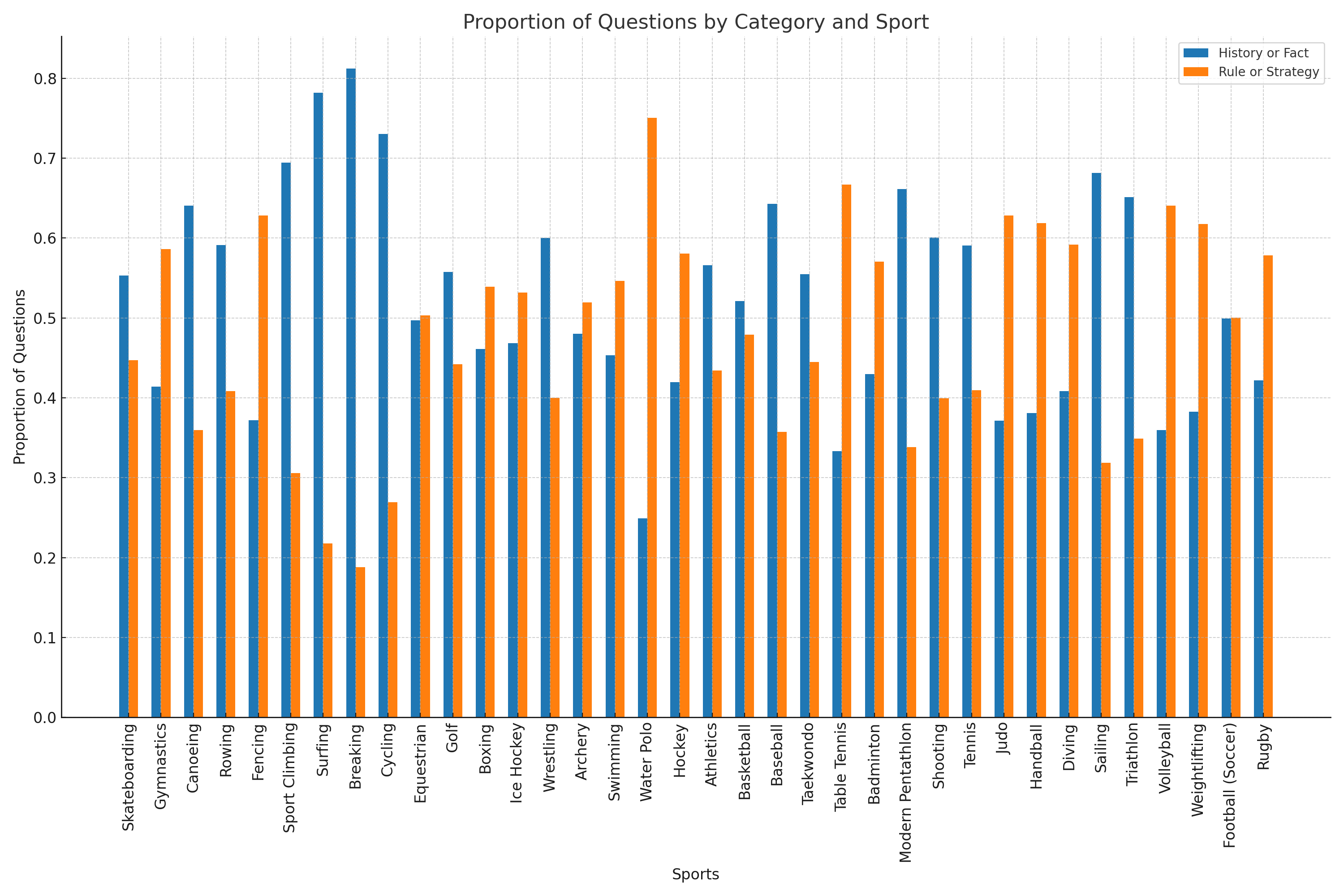}
\label{bar_chart}
\end{figure*}

A comprehensive table and corresponding figures illustrate the number of questions per sport, categorized by the type of knowledge they assess — historical facts, rules, or strategic understanding. The distribution of the number of questions between 'Rule or Strategy'  and 'History or Fact' is shown in Figure~\ref{pie_chart}. The detailed distributions for each task in level-2 are shown in Figure~\ref{bar_chart} and Table~\ref{tab:total_questions_sports}.

\begin{table}[h]
  \begin{adjustbox}{width=\columnwidth}
\centering
\begin{tabular}{lll}
\hline
\textbf{Level} & \textbf{Question Type} & \textbf{Template Examples} \\ \hline
\multirow{6}{*}{\centering level-2}&History or Fact & 
\begin{tabular}[c]{@{}l@{}}
1. In [sport], what is the notable history of [aspect]? \\
2. Who was the first in [sport] to [achievement/action]? \\
3. What is the origin of [aspect] in [sport]? \\
4. What record does [athlete/team] hold in [sport]? \\
5. Which athlete or team is known for [achievement] in [sport]?
\end{tabular} \\ \cline{2-3}
  & Rules or Strategy & 
\begin{tabular}[c]{@{}l@{}}
1. How is [technique] performed in [sport]? \\
2. What is the official rule of [aspect] in [sport]? \\
3. What strategic approach is used for [aspect] in [sport]? \\
4. Explain the play method of [situation] in [sport]. \\
5. Describe the regulation for [action/equipment] in [sport].
\end{tabular} \\ \hline
\end{tabular}
\end{adjustbox}
\caption{General templates used for generating questions in level-2 of SportQA. The templates are divided by question type (History or Fact, and Rules or Strategy) within level-2, with each category containing five examples.}
\label{template}
\end{table}

\subsection{Question Templates}
\begin{figure*}[h]
\centering
\includegraphics[width =1\textwidth]{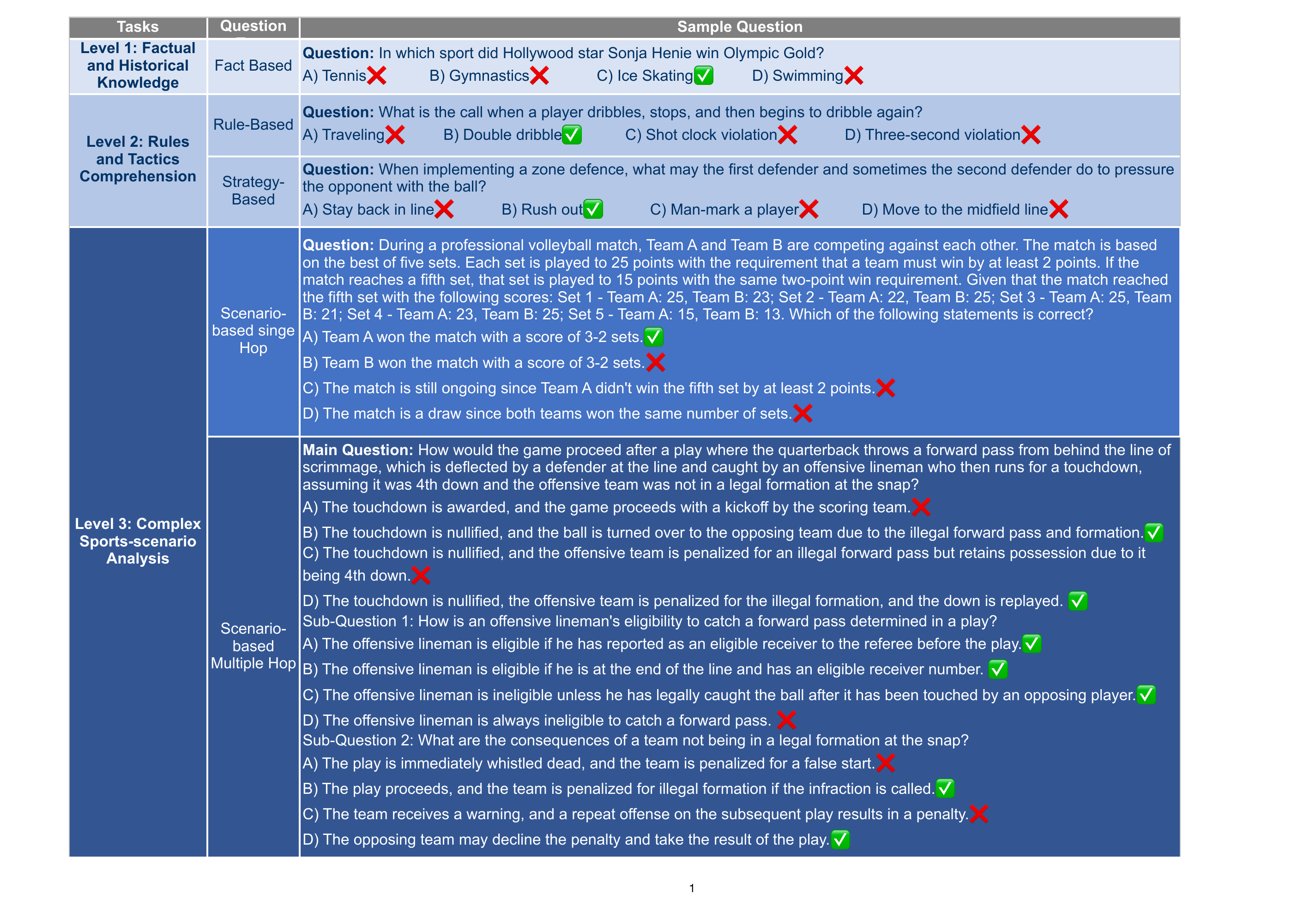}
\caption{Example questions for SportQA}
\label{framework}
\end{figure*}

\begin{table}[h]
\caption{Total Number of Questions for Each Sports Task in level-2}
\begin{adjustbox}{totalheight=0.6\textheight}
\centering
\begin{tabular}{|l|c|}
\hline
Sport & Total Questions \\
\hline
Skateboarding & 1031 \\
Gymnastics & 826 \\
Canoeing & 415 \\
Fencing & 382 \\
Sport Climbing & 72 \\
Surfing & 257 \\
Breaking & 149 \\
Cycling & 1102 \\
Equestrian & 159 \\
Golf & 2605 \\
Boxing & 1423 \\
Ice Hockey & 3025 \\
Wrestling & 170 \\
Archery & 127 \\
Swimming & 710 \\
Water Polo & 465 \\
Hockey & 417 \\
Athletics & 1398 \\
Basketball & 4400 \\
Baseball & 512 \\
Taekwondo & 182 \\
Table Tennis & 273 \\
Badminton & 149 \\
Modern Pentathlon & 65 \\
Shooting & 606 \\
Tennis & 2526 \\
Judo & 810 \\
Handball & 391 \\
Diving & 191 \\
Sailing & 430 \\
Triathlon & 172 \\
Volleyball & 1263 \\
Weightlifting & 115 \\
Football (soccer) & 8062 \\
Rugby \& American football & 10956 \\
\hline
\end{tabular}
\end{adjustbox}
\label{tab:total_questions_sports}

\end{table}

The example of templates we used to generate level-2 questions is shown in Table~\ref{template}.

\subsection{level-3 Assessment Angles}
This section presents an example of assessment angles, shown in Table~\ref{baksetball_ang}, for level-3, as proposed by sports experts. These angles are designed to challenge Large Language Models (LLMs) with realistic and complex sports scenarios that require an advanced understanding of game dynamics, rules, and strategies. For a detailed exposition of these angles within the context of basketball, please refer to Table \ref{baksetball_ang}.

\subsection{Sample Questions}

The sample questions for each level are shown in Figure~\ref{framework}. 

\section{Prompting Methods}
\label{sec:appendixB}
We utilize both standard Prompt (SP) and CoT in our experiments with LLMs. For SP, questions are presented without additional steps in the prompt. For CoT, zero-shot prompting is inspired by \citep{kojima2022large}, instructing the model to “Let's think step by step”. For few-shot CoT, we manually craft the step-by-step process for 5-shot exemplars in the development set. Examples of Few-shots and CoT prompts in our experiments are included in Table \ref{level5-sort} - \ref{tab:5cot-American}.

\section{Additional Results on GPT-4}
The GPT-4 Performance of individual tasks in level-2 and level-3 are shown in Table~\ref{tab:level-2-gpt} and Table ~\ref{tab:level3-res}. We notice 5-Shot CoT prompting has the best performance in most of the tasks. 

\begin{table}[h]
	\small
	\newcommand{\tabincell}[2]{\begin{tabular}{@{}#1@{}}#2\end{tabular}}
\centering
\begin{tabular}{cccc}
\hline
\textbf{Sports}&\textbf{\tabincell{c}{0-Shot \\CoT (\%)}}&\textbf{\tabincell{c}{5-Shot\\ SP (\%)}}&\textbf{\tabincell{c}{5-shot \\CoT (\%)}}\\\hline
Water Polo&66.67&78.26&{\bf 81.15}\\
Baseball&68.24&75&{\bf 80.26}\\
\tabincell{c}{Rugby or American Football}&69.51&{\bf 75.6}&73.17\\
Athletics&71.05&73.91&{\bf 78.26}\\
Cycling&70.37&79.62&{\bf 81.48}\\
Table Tennis&72.5&82.5&{\bf 82.56}\\
Badminton&65.11&76.74&{\bf 79.07}\\
Football (Soccer)&67.16&85.07&{\bf 85.07}\\
Sailing&61.9&74.6&{\bf 79.36}\\
Ice Hockey&66.66&{\bf 80}&76\\
Diving&51.78&76.78&{\bf 78.57}\\
Tennis&77.42&82.25&{\bf 88.7}\\
Handball&74.13&{\bf 87.93}&84.48\\
Weightlifting&84.37&87.5&{\bf 87.5}\\
Equestrian&36.95&{\bf 63.04}&50\\
Surfing&70.27&75.67&{\bf 81.08}\\
Hockey&64.51&67.74&{\bf 74.19}\\
Judo&72.5&77.5&{\bf 77.5}\\
Sport Climbing&85&95&{\bf 95}\\
Volleyball&69.35&{\bf 74.19}&72.58\\
Gymnastics&60.97&73.17&{\bf 75.6}\\
Modern Pentathlon&77.77&66.67&{\bf 94.44}\\
Taekwondo&71.69&79.24&{\bf 86.79}\\
Archery&80.55&{\bf 83.33}&80.55\\
Canoeing&80.52&72.73&{\bf 80.52}\\
Shooting&70&76.67&{\bf 84.44}\\
Wrestling&74&78&{\bf 78}\\
Boxing&71.42&{\bf 84.28}&80\\
Triathlon&72&{\bf 80}&78\\
Breaking&76.74&83.72&{\bf 86.46}\\
Golf&64.61&{\bf 84.61}&83.07\\
Basketball&69.72&71.55&{\bf 72.47}\\
Fencing&71.42&71.42&{\bf 80.35}\\
Skateboarding&60.78&60.78&{\bf 66.67}\\
Swimming&69.52&69.52&{\bf 70.47}\\
\hline
\end{tabular}
\caption{GPT-4 Performance of each task in Level-2}
\label{tab:level-2-gpt}
\end{table}

\begin{table}
	\scriptsize
	\newcommand{\tabincell}[2]{\begin{tabular}{@{}#1@{}}#2\end{tabular}}
	\centering
	\setlength{\tabcolsep}{0.1mm}
	\begin{tabular}{cccccc}
		\hline
		\textbf{Sports}&\textbf{\tabincell{c}{Prompt \\Setting}}&\textbf{\tabincell{c}{Easy \\Single-hop\\ (\%)}}&\textbf{\tabincell{c}{Hard Single-\\hop Tasks\\ (\%)}}&\textbf{\tabincell{c}{Easy Multi-\\hop Tasks \\(\%)}}&\textbf{\tabincell{c}{Multi-hop \\Tasks\\ (\%)}}\\\hline
		&Zero-shot CoT&75&56.25&39.39&21.88\\
		\tabincell{c}{American Football}&\tabincell{c}{5-Shot SP}&81.25&{\bf 71.88}&39.39&28.13\\
		&5-Shot CoT&{\bf 84.38}&68.75&{\bf 42.42}&{\bf 31.25}\\
		&Zero-shot CoT&77.42&73.33&33.87&22.95\\
		Soccer&\tabincell{c}{5-Shot SP}&80.65&78.33&37.1&24.59\\
		&5-Shot CoT&{\bf 83.87}&{\bf 81.67}&{\bf 37.1}&22.95\\
		&Zero-shot CoT&58.54&42.86&22.73&21.62\\
		Volleyball&\tabincell{c}{5-Shot SP}&73.17&57.14&20.45&18.92\\
		&5-Shot CoT&{\bf 75.61}&{\bf 59.52}&{\bf 27.27}&{\bf 21.62}\\
		&Zero-shot CoT&65.91&56.82&27.27&22.92\\
		Basketball&\tabincell{c}{5-Shot SP}&61.36&59.09&27.27&{\bf 25}\\
		&5-Shot CoT&{\bf 65.91}&{\bf 59.09}&{\bf 29.55}&20.83\\
		&Zero-shot CoT&67.86&48.15&42.86&26.92\\
		Table Tennis&\tabincell{c}{5-Shot SP}&71.43&59.26&32.14&{\bf 34.62}\\
		&5-Shot CoT&{\bf 71.43}&{\bf 62.96}&{\bf 32.14}&30.77\\
		&Zero-shot CoT&53.85&42.5&30.77&{\bf 20}\\
		Tennis&\tabincell{c}{5-Shot SP}&53.85&{\bf 47.5}&{\bf 46.15}&20\\
		&5-Shot CoT&{\bf 56.41}&45&38.46&14.29\\
		\hline
	\end{tabular}
	\caption{GPT-4 Performance of each task in Level-3}
	\label{tab:level3-res}
\end{table}

\label{sec:appendixC}

\section{Error Categories Explanation}
\label{sec:appendixD}
Examples of incorrect outputs and their corresponding error analyses are located in Table~\ref{level3-5-cot-output1} and Table~\ref{level3-5-cot-output2}.

\noindent \textbf{Level-1 and -2 error Categories.} Deficiency in Conceptual Understanding: The model demonstrates a lack of understanding or awareness of the specific sports concept or event. Misuse of Known Information: The model uses relevant information but provides incorrect explanations or conclusions or applies a general rule or concept to a specific sports scenario, leading to inaccurate answers. Incorrect Fact Recall: The model recalls and uses a fact that is either incorrect or not applicable to the given context, despite understanding the general concept. 

\noindent \textbf{level-3 error Categories.} Conceptual Misunderstanding: Errors where the model fails to grasp essential concepts required for answering the questions. Logical Reasoning Error: Mistakes in linking multiple steps of reasoning (For example, score calculation). Contextual Misinterpretation: Instances where the model misinterprets the scenario or context, leading to incorrect conclusions.

\begin{table*}[htbp]
\centering
\caption{Example of assessment angles proposed by experts: basketball assessments angles}
\begin{adjustbox}{totalheight=0.95\textheight}
\begin{tabular}{|p{0.95\textwidth}|}
\hline
Game duration, break times\\
Number of players in a team, number of substitutes\\
Jump ball rules at the start\\
Scoring rules (two-pointers, three-pointers, free throws)\\
Differentiating between personal fouls, flagrant fouls, and technical fouls\\
Determining if a player's actions, like traveling or double dribbling, are violations\\
Rules about fouling out after a certain number of personal fouls\\
Team foul limits and free throw rules\\
Number of time-outs allowed per team per quarter\\
Duration of time-outs\\
Rules and timings for making substitutions\\
Shot clock rules (24-second rule)\\
Backcourt violation and the 8-second rule\\
Defensive three-second violation\\
Determining when the ball is considered out of bounds\\
Deciding which team gets possession after the ball goes out of bounds\\
Over-and-back violation\\
Dimensions and markings of a basketball court\\
Specifications of the basketball, hoop, and backboard\\
Regulations about uniforms and footwear.\\
Dealing with interruptions or suspensions (e.g., equipment malfunction, player injuries)\\
Overtime rules\\
Handling disputes between players, coaches, and referees\\
Recording player stats like points, rebounds, assists\\
Keeping track of team stats like fouls, time-outs\\
Special rules for the last two minutes of a game\\
Ball possession rules after free throws\\
Handling technical fouls, flagrant fouls, and on-court altercations\\
How to quickly and accurately judge players' positions and actions during a game\\
How to choose the best observation point and angle based on the game's pace and players' positions\\
Differentiating between common fouls, flagrant fouls, and technical fouls\\
Judging whether physical contact between players constitutes a foul\\
Determining if a player's actions violate game rules (e.g., traveling, double dribbling)\\
Judging whether the ball has completely crossed the boundary line\\
Determining which player last touched the ball to decide ball possession\\
Ensuring the smooth progression of the game and avoiding unnecessary interruptions\\
Handling unexpected situations during a game, such as player injuries or court issues\\
Effectively communicating with players and coaches when making calls and explaining decisions\\
Handling disputes and protests from players and coaches\\
Ensuring that calls made throughout the game are based on a consistent set of standards\\
Avoiding biases or external influences on decisions\\
Managing unexpected situations like player altercations or spectators entering the court\\
Ensuring the safety and fairness of the game\\
Technical Gestures and Signals\\
Using standardized referee gestures and signals to convey decisions accurately\\
Ensuring that all participants understand and accept the calls\\
Handling disputes and complaints after the game\\
Communicating with other referees, players, and coaches for feedback and improvement\\
Positioning and Movement\\
Communication and Management\\
Mental Strength\\
Physical Fitness\\
Game Strategy and Techniques\\
Case Study Analysis\\
Ethics and Professionalism\\
Safety and First Aid \\
\hline
\end{tabular}
\end{adjustbox}
\label{baksetball_ang}
\end{table*}

\begin{table*}
\begin{tabular}{p{16cm}}
\textbf{Question:} The 1932 German football championship Final was played at a stadium that opened in what year? \\
A) 1912
B) 1928
C) 1930
D) 1932
\\ \textbf{Answer:} B \\ \hline 
\textbf{Question:} How many events did the Nordic Combined consist of in the 1948 Olympics?
A) Two events
B) Three events
C) One event
D) Four events
\\ \textbf{Answer:} C \\ \hline 
\textbf{Question:} Who succeeded David Stern to become the Commissioner of the NBA?
A) Richard Parsons
B) Adam Silver
C) Gary Bettman
D) Roger Goodell
\\ \textbf{Answer:} B \\ \hline 
\textbf{Question:} Which sport is played on a variable ground ranging from 50x100yd minimum to 100x130yd maximum? \\
A) Baseball
B) Association Football (Soccer)
C) Basketball
D) Volleyball
\\ \textbf{Answer:} B \\ \hline 
\textbf{Question:} In American Football, how many points does a touchdown score? \\
A) Six
B) Three
C) Four
D) Five
\\ \textbf{Answer:} \\ \hline
\end{tabular}
\caption{5-Shot Prompt for Few-Shot Prompting in Level-1 Experiment}
\label{level5-sort}
\end{table*}

\begin{table*}
\begin{tabular}{p{15cm}}
\textbf{Q:}  The 1932 German football championship Final was played at a stadium that opened in what year? \\
A) 1912
B) 1928
C) 1930
D) 1932 \\
\textbf{Answer:} Choices A, C, D are wrong since these years are irrelevant to the stadium where German football championship is. Hence, the correct answer is B. \\ \hline
\textbf{Q:}  How many events did the Nordic Combined consist of in the 1948 Olympics? \\
A) Two events
B) Three events
C) One event
D) Four events\\
\textbf{Answer:} In 1948 Olympics, the Nordic Combined consisted of only one event. Hence, the correct answer is C. \\ \hline
\textbf{Q:} Who succeeded David Stern to become the Commissioner of the NBA?
A) Richard Parsons
B) Adam Silver
C) Gary Bettman
D) Roger Goodell\\
\textbf{Answer:} David Stern served as the Commissioner of the NBA from 1984 until February 1, 2014. Adam Silver succeeded David Stern to become the Commissioner of the NBA. Hence, the correct answer is B. \\ \hline
\textbf{Q:} Which sport is played on a variable ground ranging from 50x100yd minimum to 100x130yd maximum?
A) Baseball
B) Association Football (Soccer)
C) Basketball
D) Volleyball\\
\textbf{Answer:} Choices A, C, D are wrong because their court size is all less the 50*100 yards. Hence, the correct answer is B. \\ \hline
\textbf{Q:} In American Football, how many points does a touchdown score? \\
A) Six
B) Three
C) Four
D) Five\\
\textbf{Answer:}Based on the fact of the rules of American football, a touchdown scores 6 points. Hence, the correct answer is A. \\ \hline
\end{tabular}
\caption{5-Shot Prompt for Few-Shot CoT Prompting in Level-1 Experiment}
\label{tab:level1_cot}
\end{table*}

\begin{table*}
\begin{tabular}{p{16cm}}
\textbf{Question:} What is the result of a bout when enough rounds have passed after an accidental foul that causes injury?   \\
A) The injured fighter loses 
B) The fight goes to a decision 
C) The fight is restarted
D) The uninjured fighter is disqualified
\\ \textbf{Answer:} B \\ \hline 
\textbf{Question:} What defensive maneuver involves rotating the body to allow an incoming punch to pass harmlessly by? \\
A) Bobbing 
B) Blocking 
C) Slipping 
D) Clinching
\\ \textbf{Answer:} C \\ \hline 
\textbf{Question:} What is the Filipino martial art that includes boxing known as? \\
A) Karate 
B) Suntukan 
C) Taekwondo 
D) Kung Fu
\\ \textbf{Answer:}B \\ \hline 
\textbf{Question:} The "door policy" at events is used to decide: \\
A) Who will perform at the event 
B) The theme of the event 
C) Who will be admitted and at what price 
D) The duration of the event
\\ \textbf{Answer:} C \\ \hline 
\textbf{Question:}  Which boxer is NOT widely recognized as one of the best bolo punchers in boxing history? \\
A) Ceferino Garcia 
B) Roy Jones Jr. 
C) Floyd Mayweather 
D) Sugar Ray Leonard
\\ \textbf{Answer:} C \\ \hline
\end{tabular}
\caption{5-Shot Prompt for Few-Shot Prompting in level-2 Boxing Task Experiment}
\label{tab:Boxing_level2}
\end{table*}

\begin{table*}
\begin{tabular}{p{15cm}}
\textbf{Q:} What is the result of a bout when enough rounds have passed after an accidental foul that causes injury?  \\
A) The injured fighter loses
B) The fight goes to a decision
C) The fight is restarted
D) The uninjured fighter is disqualified \\
\textbf{Answer:} Choices A and D are incorrect because injuries do not matter for the final fight outcome and Choice C is incorrect because boxing matches are never restarted. Hence, choice B is correct. \\ \hline
\textbf{Q:} What defensive maneuver involves rotating the body to allow an incoming punch to pass harmlessly by? \\
A) Bobbing 
B) Blocking 
C) Slipping 
D) Clinching\\
\textbf{Answer:} Slipping rotates the body slightly so that an incoming punch passes harmlessly next to the head. Hence, the correct answer is C. \\ \hline
\textbf{Q:} What is the Filipino martial art that includes boxing known as?  \\
A) Karate 
B) Suntukan 
C) Taekwondo 
D) Kung Fu\\
\textbf{Answer:} Choices A, C, and D are incorrect as these martial arts are not from the Phillipines and have very specific and distinct moves from boxing. Hence, the correct answer is B. \\ \hline
\textbf{Q:} The "door policy" at events is used to decide: \\
A) Who will perform at the event 
B) The theme of the event 
C) Who will be admitted and at what price 
D) The duration of the event\\
\textbf{A:} Choices A, B, and D are incorrect because these options do not involve general admittance. Hence the correct answer is C. \\ \hline
\textbf{Q:} Which boxer is NOT widely recognized as one of the best bolo punchers in boxing history? \\
A) Ceferino Garcia 
B) Roy Jones Jr. 
C) Floyd Mayweather 
D) Sugar Ray Leonard\\
\textbf{Answer:} Choices A, B, and D are incorrect because these three are widely known for popularizing and being excellent users of the bolo punch. Hence, the correct answer is C. \\ \hline
\end{tabular}
\caption{5-Shot Prompt for Few-Shot CoT Prompting in Level-2 Boxing Task Experiment}
\label{tab:Boxing_level2_COT}
\end{table*}

\begin{table*}
\begin{tabular}{p{16cm}}
\textbf{Main Question:}  Why might a coach call a timeout late in the fourth quarter when the team is on defense and the opposing team is on the third down?   \\
A. To rest the defensive players and prevent a score
B. To discuss strategy to force a turnover
C. To argue with the referee about a previous play
D. To allow a television commercial break\\
\textbf{Sub-Question 1:} What is the purpose of a timeout in American football?\\
A. To stop the clock
B. To make strategic adjustments
C. To replace the football
D. To challenge a referee's decision\\
\textbf{Sub-Question 2:} What can happen on a third down in American football that would prompt a defensive coach to call a timeout?\\
A) The defense needs to prevent the offense from scoring
B) The defense wants to conserve time for their offense
C) The offense is likely to punt on the next down
D) The offense might attempt a field goal\\
\textbf{Answer: }In American Football, timeouts are usually called to stop the clock, make lineup or strategy adjustments, or allow players to rest at an important moment. Hence the correct answers to the Main Question are A and B. And the correct answers to Sub-Question 1 are A and B. A pivotal third down late in the fourth quarter could make the difference between the opponent scoring a touchdown, a field goal, or nothing, so this play is very important to potentially prevent a score or decrease the amount scored by the opponent. Additionally, if a score is unavoidable or if a defensive stop is likely but your team is still trailing, a timeout on this play can save time for your team's offense to score in return and potentially take back the lead. Hence the correct answers to Sub-Question 2 are A and B.
\\ \hline 
\textbf{Main Question:} How do the responsibilities of an American Football team's offensive and defensive coordinators differ during a game?\\
A) Both design and call plays for the team
B) One focuses on scoring strategies while the other focuses on preventing the opponent from scoring
C) Both coordinate player substitutions and adjustments on their side of the ball
D) One oversees the development of the game plan, while the other implements it during the game
\\
\textbf{Sub-Question 1:} How does an offensive coordinator contribute to a team's performance in a game?
\\
A) By calling offensive plays
B) By managing the defense's tactics
C) By designing scoring strategies
D) By coaching the special teams unit
\\
\textbf{Sub-Question 2:} How does a defensive coordinator contribute to a team's performance in a game?
\\
A) By implementing the head coach's overall strategy
B) By calling defensive plays
C) By managing the offense's tactics
D) By designing strategies to prevent the opponent from scoring\\
\textbf{Answer: }The Offensive Coordinator and Defensive Coordinator in American Football both design and coordinate plays and strategies as well as decide substitutions and lineup or strategy adjustments in game for their respective sides of the ball. Hence the correct answers to the Main Question are A, B, and C. And the correct answers to Sub-Question 1 are A and C. And the correct answers to Sub-Question 2 are B and D.
\\ \hline 
\textbf{Main Question:} Why might a high school football coach enforce strict adherence to tackling techniques during practice sessions?\\
A) To ensure the team wins more games.
B) To minimize the risk of injury to players.
C) To make practices more challenging.
D) To comply with state sporting regulations.
\\
\textbf{Sub-Question 1:} What is one of the primary concerns that safe tackling techniques aim to address during football games?
\\
A) Improving player coordination.
B) Reducing the risk of concussions.
C) Enhancing the entertainment value of the game.
D) Increasing the speed of the players.\\
\textbf{Sub-Question 2:} Why are concussions a significant concern in contact sports like American football?\\
A) They can lead to long-term health issues.
B) They often require expensive equipment to diagnose.
C) They result in penalties during the game.
D) They are not detectable until days after the game.\\
\textbf{Answer: }Strict tackling techniques in American Football are typically used to prevent injuries, primarily concussions from helmet-to-helmet contact. Although sport regulations typically ban helmet-to-helmet contact, these regulations are not typically very strict about tackling form, so a coach enforcing strict tackling technique is not likely to be for following regulations. Hence the correct answer to the Main Question is A. And the correct answer to Sub-Question 1 is B. Concussions are a large concern because of the potential long-term and severe issues that can come from them. Hence the correct answer to Sub-Question 2 is A. 
\end{tabular}
%\caption{5-shot prompt for few-shot prompting in level-2 boxing task experiment}
\label{tab:5-shot-American}
\end{table*}

\begin{table*}
	\begin{tabular}{p{16cm}}
 \hline\\ 
\textbf{Main Question:} What play should the coach call next if they need to maintain possession to run down the clock at the end of the 4th quarter, considering the team is currently ahead, it's 3rd down with 2 yards to go, and the opposing team still has 2 timeouts?
\\
A) A deep passing play
B) A quarterback kneel
C) A short-yardage running play
D) A punt
\\
\textbf{Sub-Question 1:} What is the primary objective for the team that is ahead near the end of the 4th quarter?
\\
A) To score as quickly as possible
B) To maintain possession and run down the clock
C) To allow the other team to score
D) To stop the game clock
\\
\textbf{Sub-Question 2:} On a 3rd down with 2 yards to go, which type of play is most likely to achieve a first down and continue possession?
\\
A) A deep passing play
B) A quarterback kneel
C) A short-yardage running play
D) A punt\\
\textbf{Answer: }In order to run down the clock, American Football teams typically run the ball or do a quarterback kneel because incomplete passes will stop the clock. In a situation where there is a 3rd down with 2 yards to go and the opponent has two timeouts, a qb kneel will not help get a first down to continue running off the clock. Hence the correct answer to the Main Question is C. A team that is ahead will want to run down the clock and maintain their lead as best as possible. Hence the correct answer to Sub-Question 1 is B. On a 3rd down with 2 yards to go, a short run play or short pass play will be most likely to acheive a first down. Hence the correct answer to Sub-Question 2 is C.
\\ \hline 
\textbf{Main Question:} Why might a defensive coordinator instruct a linebacker to shift his positioning from a standard 4-3 defense alignment to a position directly over the opposing team's tight end before the snap in a 3rd-and-long situation?
\\
A) To better defend against a possible run play to the outside.
B) To apply pressure on the quarterback by blitzing through the gap.
C) To cover the tight end, anticipating a pass in a likely passing down.
D) To confuse the offensive line's blocking scheme.
\\
\textbf{Sub-Question 1:} Why would a defensive coordinator anticipate a pass in a 3rd-and-long situation?
\\
A) The offensive team needs to gain a significant number of yards.
B) Running plays are more effective in long-yardage situations.
C) The defense has been successful in stopping the run all game.
D) The offensive team's star running back is injured.
\\
\textbf{Sub-Question 2:} Why would a linebacker be tasked with covering the tight end instead of a defensive back in this scenario?
\\
A) The linebacker is typically faster than the defensive back.
B) The tight end is a less skilled receiver than the wide receivers.
C) The defensive back is occupied with covering a wide receiver.
D) The tight end is known for exceptional blocking rather than receiving.\\
\textbf{Answer: } In American Football, linebackers lining up directly over a player typically means they will be covering them in the case they run a passing route. To blitz or contain run plays, linebackers will usually line up over a gap, and to confuse the offensive line they will usually fake blitzes. Hence the correct answer to the Main Question is C. In a 3rd-and-long situation, the offense has a lot of space to make up to get a first down. The most likely plays to gain large yardages in a single play are deeper passing plays, though they are more risky. Hence the correct answer to Sub-Question 1 is A. Tight ends are typically check-down passing options, meaning the primary option wide receivers are usually more important to have good coverage on. Hence defensive backs will usually cover wide receivers or more important zones, leaving the less important tight end receiver to be covered by the linebacker who is not as good in pass coverage. Hence the correct answer to Sub-Question 2 is C.
\\ \hline 
\end{tabular}
\caption{5-Shot COT Prompt for Few-shot Prompting in level-3 American Football Experiment}
\label{tab:5cot-American}
\end{table*}

\begin{table*}
\begin{tabular}{p{16cm}}
\textbf{Main Question:}  How might the officiating crew determine that a double foul scenario occurred during a down where a defensive pass interference was also suspected?\\
A) By reviewing if both teams committed fouls independent of each other during the same down.
B) By assessing if the defensive team committed pass interference after the offense had already committed a holding foul.
C) By identifying if the offensive team committed a false start before the defensive player interfered with a receiver.
D) By determining if the defensive pass interference occurred simultaneously with an offensive holding.\\
\textbf{Sub-Question 1:} How can the officials confirm that the defensive team committed pass interference?\\
A) By checking if the defender made contact with the receiver before the ball was touched.
B) By determining if the defender was playing the ball and not the receiver.
C) By observing if the receiver was beyond the line of scrimmage when contact was made.
D) By ensuring that the pass was catchable and the contact impeded the receiver's opportunity to catch the ball.\\
\textbf{Sub-Question 2:} How would the officials ascertain that the offensive team committed a holding foul?\\
A) By establishing if an offensive player used his hands outside the defender's frame to restrict movement.
B) By verifying if the offensive player's hands were inside the defender's shoulders while the defender was attempting to move past.
C) By determining if the offensive player tackled the defender to the ground.
D) By observing if the offensive lineman maintained a blocking position without grabbing the defender's jersey or pads.\\
\textbf{Correct Answers: }\\
Main Question: A and B\\
Sub-Question 1: A, B, and D\\
Sub-Question 2: A, C, and D\\
\textbf{GPT-4 Answer: }In American Football, a double foul scenario occurs when both teams commit fouls independent of each other during the same down. Hence the correct answer to the Main Question is A. Defensive pass interference is confirmed when the defender makes contact with the receiver before the ball is touched, and the pass was catchable and the contact impeded the receiver's opportunity to catch the ball. Hence the correct answers to Sub-Question 1 are A and D. A holding foul by the offensive team is ascertained when an offensive player uses his hands outside the defender's frame to restrict movement. Hence the correct answer to Sub-Question 2 is A.\\
\textbf{Explanation: }\\
Main Question: GPT-4 correctly understands that separate fouls are independently reviewed, but fails to reason that in this dual foul scenario, this potential defensive pass interference could have occurred after an offensive holding penalty. This is a Logical Reasoning Error.\\
Sub-Question 1: GPT-4 fails to include the fact that incidental contact when a defender attempts to play the ball does not constitute pass interference in it's reasoning. This is a Logical Reasoning Error.\\
Sub-Question 2: GPT-4 shows clear understanding of the rules regarding offensive holding, but fails to reason that tackling a defensive player also follows this definition and that analyzing the inverse of a definition logically follows as analyzing the definition. This is a Logical Reasoning Error.\\
\hline 
\textbf{Question:} In a regular basketball game, which of the following options correctly represents the number of players in a team on the court and the number of substitutes that can be made?\\
A) 5 players on the court, 7 substitutes
B) 6 players on the court, 5 substitutes
C) 7 players on the court, 3 substitutes
D) 4 players on the court, 6 substitutes
\\
\textbf{Correct Answers: }A\\
\textbf{GPT-4 Answer: }In a regular basketball game, each team has five players on the court at a time. The number of substitutes can vary depending on the league and level of play, but there is typically no limit on the number of substitutions that can be made during a game. Therefore, none of the options provided are correct.\\
\textbf{Explanation: }\\
GPT-4 misunderstands the context of what "substitutes" are referring to. Though it is correct that a team can make infinite substitutions in a basketball game, "substitutes" in this question are referring to the substitute players a team currently has on the bench. This is a Contextual Misinterpretation Error.\\
\hline 
\caption{GPT-4 5-shot CoT Incorrect Outputs and Error Analysis}
\label{level3-5-cot-output1}
\end{tabular}
\end{table*}

\begin{table*}
\begin{tabular}{p{16cm}}
\textbf{Question:} Why might a mediator be necessary during a football match?\\
A) To ensure the players follow the rules of the game
B) To resolve disputes or conflicts between players or teams
C) To ensure the spectators are entertained
D) To keep track of the score
\\
\textbf{Correct Answers: }B\\
\textbf{GPT-4 Answer: }A mediator, in this case, the referee, is necessary during a football match to ensure the players follow the rules of the game and to resolve any disputes or conflicts that may arise between players or teams. Hence, the correct answers are A and B.\\
\textbf{Explanation: }\\
GPT-4 misunderstands the word mediator. By definition a mediator is an intermediary party in a conflict between 2 or more other parties. A referee will not need to be a "mediator" for any of their other responsibilities. This is a Conceptual Misunderstanding Error.\\
\hline 
\textbf{Question:} During the final set of a Grand Slam tennis match, Player A decided to adopt a more aggressive playing style, frequently rushing to the net after his serve. Why might Player A have made this strategic decision?\\
A) Player A noticed that Player B struggles to return low volleys effectively.
B) Player A wanted to use his strong net game to put pressure on Player B.
C) Player A was trying to conserve energy by reducing the length of the rallies.
D) Player A was attempting to intimidate Player B by showing off his physical prowess.
\\
\textbf{Correct Answers: }A, B, and C\\
\textbf{GPT-4 Answer: }In tennis, a player might adopt a more aggressive playing style and frequently rush to the net after serving to put pressure on the opponent, exploit the opponent's weakness in returning low volleys, reduce the length of the rallies to conserve energy, or to intimidate the opponent. Hence the correct answers are A, B, C, and D.\\
\textbf{Explanation: }\\
GPT-4 either misunderstands why an aggressive playing style is used in tennis or confuses the concepts of intimidation and applying pressure to an opponent. In non-contact sports such as tennis, aggressive playstyles are typically not used for intimidation, though they are frequently used to put pressure on an opponent to make them cause errors or break down mentally. Thus GPT-4 may be confusing the concepts of intimidation and applying pressure to cause the opponent to make errors. This is a Conceptual Misunderstanding Error.\\
\hline 
\end{tabular}
\caption{GPT-4 5-shot CoT Incorrect Outputs and Error Analysis}
\label{level3-5-cot-output2}
\end{table*}

\end{document}